\documentclass[11pt]{article}

\usepackage[preprint]{acl}

\usepackage{times}
\usepackage{latexsym}
 \usepackage{amsmath}
 \usepackage[most]{tcolorbox}
 \newtcolorbox{chatbox}{
  colback=gray!5,
  colframe=gray!40,
  boxrule=0.5pt,
  arc=3pt,
  left=4pt,
  right=4pt,
  top=4pt,
  bottom=4pt,
  fontupper=\ttfamily,
}
\usepackage{booktabs,tabularx,ragged2e,enumitem,longtable}
\newcolumntype{Y}{>{\RaggedRight\arraybackslash}X}

\usepackage[T1]{fontenc}

\usepackage[utf8]{inputenc}

\usepackage{microtype}

\usepackage{inconsolata}

\usepackage{graphicx}
\usepackage{subcaption}

\title{
Word Clouds as Common Voices: LLM-Assisted Visualization of Participant-Weighted Themes in Qualitative Interviews
 }

\author{Joseph T. Colonel \\
\small Icahn School of Medicine at Mount Sinai \\
  \texttt{joseph.colonel@mssm.edu} \\\And
  Baihan Lin$^*$ \\
\small Icahn School of Medicine at Mount Sinai \\
  \texttt{baihan.lin@mssm.edu} \\}

\begin{document}
\maketitle
\begin{abstract}
Word clouds are a common way to summarize qualitative interviews, yet traditional frequency-based methods often fail in conversational contexts: they surface filler words, ignore paraphrase, and fragment semantically related ideas. This limits their usefulness in early-stage analysis, when researchers need fast, interpretable overviews of what participant actually said. We introduce \textbf{ThemeClouds}, an open-source visualization tool that uses large language models (LLMs) to generate \emph{thematic, participant-weighted} word clouds from dialogue transcripts. The system prompts an LLM to identify concept-level themes across a corpus and then counts how many unique participants mention each topic, yielding a visualization grounded in \emph{breadth of mention} rather than raw term frequency. Researchers can customize prompts and visualization parameters, providing transparency and control. Using interviews from a user study comparing five recording-device configurations (31 participants; 155 transcripts, Whisper ASR), our approach surfaces more actionable device concerns than frequency clouds and topic-modeling baselines (e.g., LDA, BERTopic). We discuss design trade-offs for integrating LLM assistance into qualitative workflows, implications for interpretability and researcher agency, and opportunities for interactive analyses such as per-condition contrasts (``diff clouds'').
\end{abstract}

\section{Introduction}

Qualitative interviews are a cornerstone of HCI practice: they capture lived experience, tacit knowledge, and situated rationales that are difficult to elicit through logs or lab tasks alone \cite{hopf2004qualitative}. But precisely because conversational data are rich, early-stage sensemaking can be slow and brittle. Time-constrained teams often rely on word clouds to orient themselves and to communicate initial patterns. Word clouds help researchers surface recurring terms and communicate high-level themes to stakeholders \cite{khusro2021tag}. In principle, a quick visualization that ``shows what people talked about'' is invaluable. In practice, however, frequency-based word clouds tend to reflect \emph{how} people talk rather than \emph{what} they mean.

This misalignment is acute for spoken transcripts. Even with stop-word removal, the statistical surface of talk often dominates frequency ranks, such as disfluencies (``uh''), discourse markers (``like'', ``you know''), and coordination (``and''). Moreover, participants rarely reuse identical strings when describing similar concerns. One person may say ``it felt in the way,'' another ``kind of distracting,'' another ``I kept noticing the device,'' and a fourth ``it made me self-conscious.'' Traditional clouds fragment these into separate tokens, spreading salience thinly across synonyms and paraphrases. The resulting picture understates a theme’s breadth and overstates lexical quirks, leaving analysts to manually reconcile meaning after the fact.

In our motivating study, clinicians and participants evaluated different recording-device configurations intended for psychiatric assessment. When we generated standard frequency clouds per device, familiar problems reappeared: conversational scaffolding rose to the top; multi-word concerns broke into stems; semantically aligned reactions (e.g., ``distracting,'' ``in the way,'' ``felt watched'') were scattered. The clouds neither matched researcher notes nor helped communicate trade-offs to stakeholders. A different aggregation principle was needed.

Recent advancements in large language models (LLMs) present new opportunities for enhancing qualitative analysis \cite{xu2025tama}. Models such as Llama 3.3 can process long passages of unstructured text, identify latent topics, and recognize semantically important terms even when they are phrased differently across transcripts \cite{touvron2023llama}. These capabilities make LLMs well-suited for tasks like summarization and topic extraction, which are core components of qualitative synthesis. 

In our use case, LLMs make a new design space feasible. Rather than counting words, we can ask a model to reason about concepts, recognize paraphrases, and collapse near-synonyms—capabilities that have matured as models improved long-context understanding. But naively inserting LLMs can reduce transparency. Our design goal, therefore, is to preserve the \emph{immediacy and communicability} of word clouds while shifting the unit of analysis from tokens to \emph{concepts}, and the weighting from raw counts to the \emph{breadth of mention} across participants. In effect, we want a cloud that answers the question analysts and stakeholders actually ask: ``How many people brought this up?''

We contribute a method and artifact that operationalize this shift in a way that fits qualitative workflows. Our open-source tool, ThemeClouds, leverages Llama 3.3 to assist in generating semantic word clouds from qualitative interview transcripts. Rather than relying solely on term frequency, the tool uses LLM reasoning to extract salient terms and conceptually related groupings, producing visualizations that better reflect the themes embedded in natural dialogue. By incorporating lightweight user control, the system balances LLM assistance with researcher agency, supporting interpretation while preserving transparency and flexibility.

Our work builds on prior literature in textual visualization and qualitative coding tools \cite{bateman2008seeing, lennon2021developing}. While previous approaches have highlighted the risks of misleading word clouds or opaque model outputs, we aim to demonstrate how thoughtful design centered around customization and interpretability can help researchers co-construct word clouds with LLMs in qualitative workflows. The remainder of this paper describes the architecture and design decisions behind the system, demonstrates its application to interview data, and reflects on broader implications for LLM-assisted tools in qualitative analysis.

Our contribution is methodological and pragmatic. We do not claim a new theory of qualitative analysis; instead, we provide a lightweight, defensible, and \emph{participant-weighted} alternative to frequency clouds that better aligns early-stage summaries with how analysts reason and report. We show how to integrate LLM assistance without obscuring the analytic process, emphasizing controls, artifacts, and audit trails that allow researchers to trust, contest, and adapt outputs.

\section{Related Work}

\subsection{Word clouds as communicative summaries}
Word (or tag) clouds have enduring appeal because they compress large corpora into a glanceable visual summary, where word frequency maps to font size. Early tools like Wordle made word clouds ubiquitous on the web \cite{steele2010beautiful}. Kaser and Lemire formalized the layout problem, showing how to use 2D packing and typesetting techniques to draw tag clouds efficiently \cite{barth2014semantic}. Subsequent work  evaluated how visual features affect readability and selection \cite{rivadeneira2007getting,bateman2008seeing}. As a result, classic word clouds can be ``aesthetically pleasing'' and easy to create 
but have well-documented limitations for analytic tasks.

These efforts improved the communicative surface, yet the core statistic -- token frequency -- remains brittle in conversational settings, where disfluency and paraphrase are the norm. Our approach retains the familiar word-cloud form while changing the underlying weighting to reflect population-level salience.

\subsection{Speech-derived clouds and semantic grouping}

Spoken language transcripts differ markedly from traditional text sources like news articles or reviews as they are spontaneous, unedited, and often noisy. Disfluencies such as filler words ("um", "like"), false starts, and repetition are commonplace. The transcript format introduces both unique structure (turn-taking, repair, backchannels) and noise (ASR errors, fillers). These properties challenge the direct application of word cloud techniques developed for clean, edited corpora. Prior work in visualization, natural language processing (NLP), and accessibility has begun addressing these issues, especially in the context of spoken interactions.

Several systems have explored real-time word cloud generation from speech. Iijima et al. designed an interface for deaf and hard-of-hearing users that visualizes each speaker’s utterances as personalized word clouds, enabling better topic tracking in meetings \cite{iijima2021word}. Importantly, their system filters out non-content words, addressing the prevalence of noise in speech. Chandrasegaran et al. similarly integrate ASR with word clouds in TalkTraces \cite{chandrasegaran2019talktraces}, emphasizing that when enhanced with topic modeling and embedding-based filtering, word clouds can help users follow evolving spoken discussions. These works highlight the value of preprocessing speech transcripts to improve word cloud clarity.

The semantic structure of speech also requires more than frequency-based layouts. Wang et al. proposed ReCloud \cite{wang2020recloud}, which clusters semantically similar terms using NLP techniques, allowing users to grasp themes rather than isolated keywords. Skeppstedt et al. extended this idea with Word Rain \cite{skeppstedt2024word}, embedding word semantics along a visual axis and combining font size with TF-IDF bar charts. Though both methods were tested on written corpora (reviews, climate texts), they underscore how semantic grouping and de-biasing frequency are crucial for domains where redundancy and ambiguity are common.



Together, these studies suggest that effective word cloud generation from speech transcripts must account for semantic ambiguity and high noise levels. This motivates approaches that combine filtering for content-bearing terms and semantically aware tags to produce meaningful visualizations of conversational speech. Our method builds on this trajectory by externalizing grouping decisions to an LLM while preserving analyst control over prompts, topic cardinality, and the final mapping.

\subsection{LLM-assisted thematic analysis}

LLMs have been used to accelerate theme discovery, propose candidate codes, and reduce analytic burden, sometimes reaching near-human agreement in semi-structured settings. They enable scalable and semi-automated approaches to thematic analysis of qualitative interviews, especially in domains where manual coding is labor-intensive. In the biomedical context, Xu et al. introduced TAMA \cite{xu2025tama}, a multi-agent LLM framework designed to assist clinicians in analyzing interviews related to congenital heart disease. By integrating human-in-the-loop feedback with AI-generated theme suggestions, TAMA enhances the accuracy and distinctiveness of identified themes, while significantly reducing the burden on expert coders. Similarly, Singh et al. developed RACER \cite{singh2024racer}, an LLM-powered methodology applied to semi-structured interviews conducted during the COVID-19 pandemic. RACER achieved near-human agreement in theme extraction, demonstrating that LLMs can reliably support mental health research involving large volumes of qualitative data.

These successes suggest that concept-level reasoning over long documents is feasible. Our contribution is to harness these capabilities for a narrow but ubiquitous task (first-pass summarization via word clouds) while foregrounding human-centered properties (agency, transparency, workflow fit) that determine whether such tools are practically useful in HCI contexts.


\section{Methods}

\begin{figure*}
    \centering
    \includegraphics[width=\linewidth]{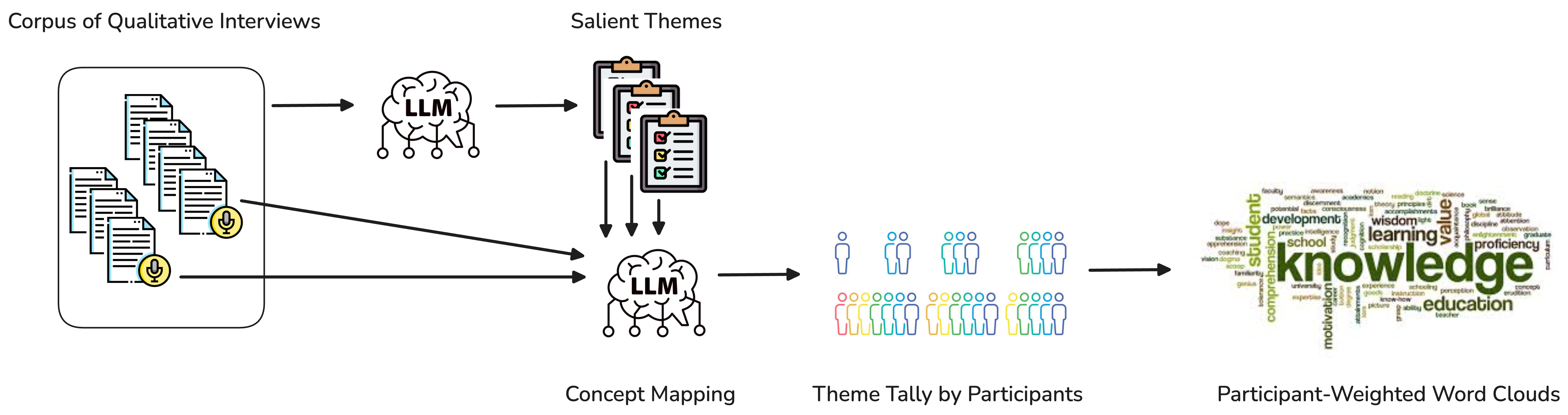}
    \caption{System overview for \textbf{ThemeClouds}: LLM-assisted \emph{participant-weighted} thematic word clouds. An LLM first proposes a compact set of concept-level themes for the corpus. Each transcript is then mapped to this fixed theme list via binary presence judgments, yielding a per-theme count of \emph{unique participants} (breadth). The final cloud sizes each theme by its participant prevalence (not token frequency). Prompts, per-transcript assignments, and counts form an audit trail that supports iteration and reproducibility.}
    \label{fig:blockdiagram}
\end{figure*}

ThemeClouds is designed to assist researchers in generating word clouds from qualitative interview transcripts by using LLMs to surface salient, semantically meaningful concepts, rather than relying on surface-level word frequency. The pipeline consists of three key stages: (1) identifying candidate concepts across a corpus, (2) mapping those concepts to individual transcripts, and (3) aggregating the results to produce a word cloud visualization. Our system prioritizes topic relevance, clarity, and interpretability over lexical frequency or length. Figure \ref{fig:blockdiagram} outlines the proposed workflow.

We formalize the shift from tokens to concepts and from frequency to breadth. Let $\mathcal{T}=\{t_1,\dots,t_M\}$ be transcripts (one per participant for a given condition) and let $\mathcal{C}=\{c_1,\dots,c_N\}$ be short concept-phrases proposed by an LLM for the corpus. For each transcript $t$ and concept $c$, the mapping step produces a binary assignment $y(t,c)\in\{0,1\}$ indicating whether the concept is clearly present in the transcript (the artifact optionally supports a soft score $\hat{p}(t,c)\in[0,1]$ with threshold $\tau$ for binarization). The \emph{breadth} of concept $c$ is:
\[
b(c) \;=\; \sum_{t\in\mathcal{T}} y(t,c),
\]
the number of unique participants whose transcripts include the concept. The visual weight for $c$ is $w(c)=g(b(c))$, where $g(\cdot)$ is a monotone scaling (linear by default; logarithmic and square-root options aid mid-rank legibility). We also support condition-wise contrasts by rendering $\Delta b(c)=b_{\text{A}}(c)-b_{\text{B}}(c)$ to make differences across device configurations glanceable.

\subsection{Input and preprocessing}

The system takes as input a collection of textual transcripts from qualitative interviews. These transcripts may come from usability studies, field interviews, focus groups, or other open-ended sources. Transcripts are assumed to be minimally cleaned (e.g., anonymized and transcribed verbatim) but do not require pre-coding or structuring. Because the method abstracts above tokens, we found that aggressive lexical normalization is unnecessary; we keep punctuation and stop-words intact for the LLM stage, using standard tooling like NLTK only for baseline clouds \cite{bird2006nltk}. Interviews are transcribed with Whisper \cite{radford2023robust}.

\subsection{Concept elicitation (corpus-level)}

The goal is a compact, human-interpretable vocabulary that captures salient ideas without collapsing distinct concerns. We prompt a long-context LLM with the corpus (or stratified subsets) to propose $N$ short concept-phrases, encouraging specificity (e.g. ``in the way,'' ``felt watched,'' ``image quality''), discouraging generic terms (``user,'' ``good,'' ``bad''), and avoiding fillers or study-task scaffolding. Rather than returning frequent unigrams or bigrams, the model is guided via prompt engineering to prioritize short phrases, semantically specific topics, and coverage diversity across the corpus. 
We favor a diverse set that covers the thematic space rather than a large list that risks redundancy. The artifact includes our exact prompts and a small set of variations. Analysts can re-run this step to explore granularity. We also explicitly discourage the model from selecting filler words, generic terms like ``user'' or ``system,'' or concepts that appear frequently but lack thematic depth. 

In our evaluation, we prompt a poplar open-source LLM model, LLaMa-3.3-70B-Instruct \cite{touvron2023llama}, to identify a set of $N$ salient topics that best represent key concepts across the entire corpus with the following prompt. 

\begin{chatbox}
You are analyzing interview transcripts where participants were asked to share their experiences using five webcam setups: [insta], [single iphone], [dual iphones], [logitech], and [obsbot].

The transcripts are organized in the following format:
Each section begins with the webcam label (e.g., "\#\#\# insta") followed by participant comments about that device.

Ignore filler words, repeated question prompts, or interviewer language. Focus only on participant speech that offers insight, reaction, or description.

Your task is to identify **exactly 20 meaningful and distinctive words or short phrases** that summarize participants’ real experiences for **each webcam setup**.

Guidelines:

- Do NOT just pick the most frequent words.

- Select words or short phrases that are **emotionally descriptive**, **technically relevant**, or **highlight distinctive qualities** (positive or negative).

- Avoid: generic words (e.g., “thing”, “camera”), filler words, or phrases repeated from the question.

For each setup, return a bullet list of 20 high-quality descriptors.

Output format:

\#\#\# [setup]
- ...
\end{chatbox}

The result is a curated list of $N$ topics that act as candidate entries for the word cloud. These phrases serve as a proxy for the major themes in the interviews, as judged by the LLM in context.

\subsection{Concept mapping (per transcript)}

In the second stage, the LLM is prompted to evaluate each transcript individually in relation to the N identified concepts or insights. For each transcript, the model receives: (1) the full content of that single transcript and (2) the fixed list of $N$ topics produced in the prior step.

The model is then tasked with identifying which topics are clearly present in the given transcript. Importantly, the prompt encourages the model to make binary or categorical judgments rather than assigning soft weights or scores. This helps mitigate overfitting and keeps results interpretable for the end user. We use the following prompt:

\begin{chatbox}
You are analyzing a participant's response about the **{device\_name}** webcam setup. 

Below is a list of key descriptive terms and phrases that were identified across interviews for this webcam. Your task is to determine **which (if any)** of these words or phrases are meaningfully reflected in the participant's comments — even if the exact wording is not used.

Focus on semantic alignment: if a participant implies or clearly expresses a concept that corresponds to one of the key terms, include it.

\#\#\# Key Descriptive Terms for **{device\_name}**:**{keyword\_list}**

\#\#\# Output Instructions:

Return ONLY a list of matching terms (one per line).

Do not include explanations, numbering, bullet points, or extra commentary.

A maximum of 20 key descriptive terms and phrases are allowed.

It is imperative to avoid false positives: if a keyword isn't reasonably supported, do not include it.
\end{chatbox}

Through the above approach, given $\mathcal{C}$, we map each transcript independently by asking the LLM to judge concept presence using the fixed vocabulary. We default to binary assignments to keep outputs interpretable and to avoid length confounds: loquacious speakers should not inflate weights. Binary judgments also simplify spot-checks: analysts can audit questionable assignments by reading short excerpts of the transcript. The artifact includes an optional soft scoring mode ($\hat{p}(t,c)$) and guidance for threshold selection if analysts prefer graded presence.

This process is repeated for every transcript in the corpus. For each topic, we then compute a relative count of the number of transcripts in which the topic was marked as present. This produces a simple but robust measure of topic salience across the corpus.
















\begin{figure*}[!th]
  \centering
  \subfloat[Frequency-based word cloud]{\includegraphics[width=0.45\textwidth]{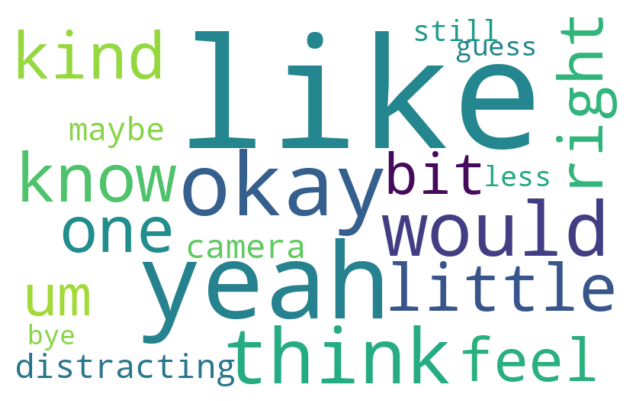}\label{fig:f1}}
  \hfill
  \subfloat[LLM-assisted word cloud]{\includegraphics[width=0.45\textwidth]{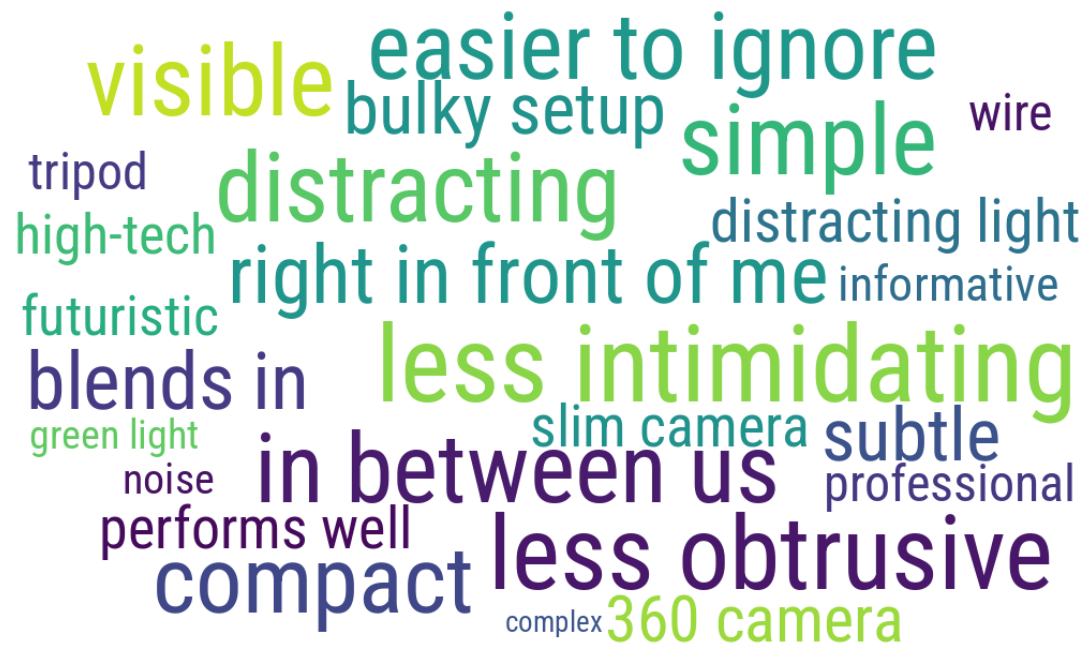}\label{fig:f2}}
  \caption{
  Side-by-side comparison for one device condition (top 20 items shown). \textbf{(a)} A traditional frequency cloud—even after stop-word filtering—elevates conversational surface tokens and fragments paraphrases. \textbf{(b)} Our LLM-assisted, participant-weighted ThemeClouds collapses paraphrases into themes and sizes each by the number of unique participants who mentioned it, foregrounding actionable concerns from the interviews.}\label{fig:wordcloud}
\end{figure*}

\begin{table*}[t]
\centering
\caption{Comparison of outputs from BERTopic, LDA, and our participant-weighted thematic method on the interview corpus. Lists are reproduced from model outputs (verbatim) and our curated themes (top items).}
\label{tab:compare}
\begin{tabularx}{\textwidth}{YYY}
\toprule
\textbf{BERTopic (Top Topics)} & \textbf{LDA (Top Topics)} & \textbf{ThemeClouds} \\
\midrule
\begin{minipage}[t]{\linewidth}
\footnotesize
\begin{enumerate}[leftmargin=*,itemsep=1pt,topsep=0pt]
\item yeah, like, maybe, okay, whatever, aware, still, um, issue, course
\item like, part, um, things, process, always, treatment, cause, really, way
\item definitely, would, bit, oh, uncomfortable, good, odd, want, fact, especially
\item yeah, like, maybe, okay, whatever, aware, still, um, issue, course
\item okay, little, look, think, least, light, um, blends, bright, get
\item yeah, even, white, side, either, light, much, like, slightly, around
\item okay, little, look, think, least, light, um, blends, bright, get
\item  definitely, would, bit, oh, uncomfortable, good, odd, want, fact, especially
\item yeah, even, white, side, either, light, much, like, slightly, around
\item like, part, um, things, process, always, treatment, cause, really, way
\end{enumerate}
\vspace{0.25em}
\end{minipage}
&
\begin{minipage}[t]{\linewidth}
\footnotesize
\begin{enumerate}[leftmargin=*,itemsep=1pt,topsep=0pt]
\item 0.005*like + 0.004*okay + 0.004*think + 0.003*would + 0.003*little
\item 0.018*like + 0.008*okay + 0.007*yeah + 0.007*think + 0.007*little
\item 0.004*like + 0.003*um + 0.003*part + 0.003*would + 0.003*things
\item 0.004*like + 0.004*okay + 0.003*think + 0.003*little + 0.003*um
\item 0.099*like + 0.034*little + 0.027*think + 0.026*okay + 0.025*bit
\item 0.139*like + 0.045*yeah + 0.027*um + 0.027*okay + 0.022*think
\item 0.174*like + 0.037*um + 0.023*things + 0.020*would + 0.020*part
\item 0.103*like + 0.057*would + 0.036*bit + 0.031*definitely + 0.026*think
\item 0.130*like + 0.029*um + 0.024*yeah + 0.018*would + 0.018*okay
\item 0.133*like + 0.046*okay + 0.046*little + 0.040*um + 0.040*think
\end{enumerate}
\vspace{0.25em}
\end{minipage}
&
\begin{minipage}[t]{\linewidth}
\footnotesize
\begin{enumerate}[leftmargin=*,itemsep=1pt,topsep=0pt]
\item Small and compact
\item Not distracting
\item Easy to ignore
\item Less noticeable
\item Not too visible
\item Fades into the background
\item Simple and straightforward
\item Convenient
\item Reminds me of a Polaroid
\item Compact and spacious
\end{enumerate}
\vspace{0.25em}
\end{minipage}
\\
\bottomrule
\end{tabularx}
\end{table*}

\subsection{Visualization and contrasts}

The final step uses these tallied topic counts to construct a word cloud. We render a conventional word cloud where size encodes $w(c)$. Each of the $N$ topics or concepts to be highlighted is included. Because the units are now people, font size directly communicates population-level salience: often the most defensible signal when communicating with product teams or clinical stakeholders. In another word, the font size of each phrase is scaled based on how frequently it was mentioned across the subjects recruited for the qualitative interviews. Terms that were mentioned in most or all transcripts are rendered largest, while rare or marginal topics appear smaller.

For comparative analysis, we can also produce condition-wise ``diff clouds'' by coloring or separating concepts whose $\Delta b(c)$ exceeds a small margin. This reveals what a device configuration uniquely amplifies or suppresses.

\subsection{Analyst-in-the-loop workflow}

A central design goal is \emph{researcher agency}. The system includes controls for adjusting the number of topics or concepts to note, the word cloud layout, font scaling, and prompt variants. Analysts can (1) edit the prompt, (2) adjust $N$, (3) seed or pin concepts they care about, (4) re-run elicitation to split overly broad concepts, and (5) audit and correct per-transcript assignments. This allows researchers to explore different perspectives on their data while retaining interpretability and structure. While the LLM outputs are fixed per run, users can rerun the topic generation with new prompts or adjusted constraints to suit different analytic goals.

We persist an \emph{assignment table} with rows as transcripts and columns as concepts so that any cloud can be reconstructed, inspected, or exported to downstream thematic coding. This audit trail helps teams defend qualitative findings in mixed-methods reports.

\section{Human-Centered Design Considerations}

A tool succeeds in HCI not only by being accurate, but by fitting how people actually work. We therefore prioritized five properties. 

\begin{itemize}
    \item \emph{Interpretability:} counting unique participants aligns with how analysts argue salience (“many people brought this up”). 
\item \emph{Transparency:} we expose prompts, concept lists, assignment tables, and scaling choices, making it easy to reconstruct decisions or contest them. 
\item \emph{Agency:} analysts can tune granularity and re-run steps to explore alternative framings. 
\item \emph{Frugality:} default settings work on small, noisy corpora typical of interviews, without heavy parameter sweeps. 
\item \emph{Workflow fit:} outputs are designed to triage and guide subsequent coding, not to replace careful qualitative analysis, echoing prior HCI work on semantic grouping and hybrid visual summaries \cite{iijima2021word,wang2020recloud,chandrasegaran2019talktraces,skeppstedt2024word}.
\end{itemize}

\section{Qualitative Evaluation}

To assess the utility of our approach in a real-world setting, we applied it to a set of qualitative interviews conducted as part of a clinical psychology research study. 31 participants evaluated five webcam setups for psychiatric outpatient clinical assessments, producing 155 interviews with a clinical research coordinator. These conversations were conducted as one-hour in-person session, in a naturalistic dialogue format, as participants and the clinical research coordinator collaboratively evaluated different hardware configurations. Audio was transcribed with Whisper \cite{radford2023robust}. 

For a representative device condition (31 transcripts), Figure \ref{fig:wordcloud} compares a standard frequency-based word cloud with NLTK stop-word removal \cite{bird2006nltk} (a) with our LLM-assisted word cloud (b). Despite identical source data, the frequency cloud elevates general discourse terms and fragments multi-word concerns, while the LLM cloud foregrounds concrete, device-specific ideas consistent with researcher notes.

\emph{What the numbers mean.} Because our weights are counts of unique participants, the magnitude of a label directly translates to breadth. If ``distracting'' appears in 20 of 31 transcripts for a device, its visual prominence is immediately defensible—helpful for design reviews and IRB or clinical discussions where conservative, population-grounded claims are preferred.

To situate the approach among common baselines, we also trained topic models such as LDA and BERTopic \cite{vrehuuvrek2011gensim,grootendorst2022bertopic,lin2023neural} on the full 155-document corpus. As in Table \ref{tab:compare}, given the small per-condition sample size and conversational style, neither produced immediately legible, per-device themes without additional manual massaging. Our participant-weighted list, on the other hand, aligns closely with analyst field notes and per-device concerns recorded during the study, foregrounding concept-level themes (e.g., ``Not distracting,'' ``Discreet,'' ``Blends into the desk'') that multiple participants independently raised.

While these observations are not a controlled user study, they illustrate a pattern we frequently saw during analysis: people-weighted concept clouds provide a more faithful ``first glance'' at what mattered to participants than token frequency or off-the-shelf topic models in this setting. It can effectively support researchers in identifying salient themes from conversational transcripts, even without structured codes or annotations.


\section{Discussion and Limitations}

Our tool demonstrates how large language models can be leveraged to assist in synthesizing qualitative feedback through semantic word clouds, offering an accessible, low-overhead entry point into exploratory analysis. While initial use cases show alignment with human interpretation, there are important limitations to consider. 

\subsection{Validity, bias, and controllability}
LLM judgments depend on prompts and may overgeneralize. The system relies on static prompts and single-pass outputs, which may overlook nuances or misrepresent concepts without user intervention. We mitigate this by using a fixed vocabulary (reducing drift), binary mapping (reducing verbosity bias), and an assignment table that supports spot-checks and corrections. Analysts can also seed concepts to ensure coverage of domain-critical concerns, an approach compatible with standard qualitative rigor practices.

\subsection{Granularity and concept drift}
The right granularity is contextual. Collapsing all camera-related concerns might hide distinctions between ``felt watched'' and ``image quality.'' While prompt customization provides some control, more interactive or iterative workflows could better support researchers in refining outputs over time. Our workflow treats concept elicitation as an iterative process: split or merge concepts, re-run mapping, and compare clouds. We found small $N$ (e.g., 12–25) balanced coverage and legibility, but analysts can tune $N$ to their corpus.

\subsection{Generalizability and small-data regimes}
The method targets the small, noisy corpora typical of interviews and focus groups. Unlike topic models, which may prefer longer documents or larger datasets, our mapping step scales down: it asks a concrete question of each transcript with a fixed vocabulary. This makes the method robust when $M$ is modest and concepts are grounded in context of the study and clinical application.

\subsection{Ethics, privacy, and deployment}
Interviews often contain sensitive information. Our artifact documents de-identification assumptions and supports local or compliant deployment. We view LLM assistance as a \emph{scaffold} for human analysis, not a replacement: analysts should verify sensitive claims and avoid over-reliance on automated judgments in consequential settings.

We position people-weighted semantic clouds as a first-pass \emph{orientation} tool. They help teams see what many participants noticed, seed codebooks, and communicate trade-offs. They do not obviate careful reading, synthesis, or theory-building. This stance aligns with prior HCI work that treats semantic grouping and hybrid visual encodings as aids to human reasoning rather than endpoints.

\subsection{Interactivity and explanation}
Static clouds are useful, but interactive affordances (such as hovering to see exemplar quotes, clicking to open transcripts, showing per-condition contrasts, toggling scaling) can turn the cloud into a navigational entry point for analysis. Because we persist per-transcript assignments, simple linkages suffice. We leave richer explanation (minimal rationales for concept presence) as future work consistent with analyst agency \cite{iijima2021word,wang2020recloud,chandrasegaran2019talktraces,skeppstedt2024word}.

Future work will focus on improving model transparency, allowing users to inspect why certain phrases were chosen or how decisions were made at the transcript level, for instance in clinical decision support tools such as \cite{lin2023psychotherapy,lin2023supervisorbot,lin2025compass}. We are also exploring ways to incorporate multi-turn refinement and lightweight feedback mechanisms, enabling more dynamic human-LLM collaboration. In parallel, more formal evaluations across domains and user roles will be important to assess the tool’s effectiveness, trustworthiness, and usability in varied qualitative research contexts.

\section{Artifact}
Our open-source ThemeClouds package \footnote{\url{https://github.com/linlab/ThemeClouds}} includes: (1) prompt templates for concept elicitation and per-transcript mapping; (2) scripts to reproduce Figure \ref{fig:wordcloud}; and (3) anonymized assignment tables and per-concept participant counts suitable for auditing and alternative visualizations. The artifact also documents default parameters and prompt variants, so other researchers can reproduce and adapt the pipeline without brittle prompt hacking. We hope this work encourages further exploration into how LLMs can provide insight in qualitative workflows.

\section{Conclusion}
We introduced ThemeClouds, a participant-weighted, concept-level approach to word clouds using LLMs to count \emph{who} raised \emph{which} ideas, aligning early-stage summaries with the way HCI and UX analysts argue salience. In an audiovisual (AV) study for clinical assessment, the method surfaced actionable concerns that frequency clouds and topic-modeling baselines obscured. By emphasizing transparency, agency, and auditability, it bridges NLP advances and qualitative practice, offering a pragmatic step toward interactive, human-centered, LLM-assisted analysis. 

\section*{Acknowledgments}
We thank the participants and research staff who made this study possible, and colleagues who provided feedback during development.
This work is supported by NIH grant 1U01MH136535.

\newpage

\bibliography{main}

\begin{thebibliography}{22}
\providecommand{\natexlab}[1]{#1}

\bibitem[{Barth et~al.(2014)Barth, Fabrikant, Kobourov, Lubiw, N{\"o}llenburg, Okamoto, Pupyrev, Squarcella, Ueckerdt, and Wolff}]{barth2014semantic}
Lukas Barth, Sara~Irina Fabrikant, Stephen~G Kobourov, Anna Lubiw, Martin N{\"o}llenburg, Yoshio Okamoto, Sergey Pupyrev, Claudio Squarcella, Torsten Ueckerdt, and Alexander Wolff. 2014.
\newblock Semantic word cloud representations: Hardness and approximation algorithms.
\newblock In \emph{Latin American Symposium on Theoretical Informatics}, pages 514--525. Springer.

\bibitem[{Bateman et~al.(2008)Bateman, Gutwin, and Nacenta}]{bateman2008seeing}
Scott Bateman, Carl Gutwin, and Miguel Nacenta. 2008.
\newblock Seeing things in the clouds: the effect of visual features on tag cloud selections.
\newblock In \emph{Proceedings of the nineteenth ACM conference on Hypertext and hypermedia}, pages 193--202.

\bibitem[{Bird(2006)}]{bird2006nltk}
Steven Bird. 2006.
\newblock Nltk: the natural language toolkit.
\newblock In \emph{Proceedings of the COLING/ACL 2006 interactive presentation sessions}, pages 69--72.

\bibitem[{Chandrasegaran et~al.(2019)Chandrasegaran, Bryan, Shidara, Chuang, and Ma}]{chandrasegaran2019talktraces}
Senthil Chandrasegaran, Chris Bryan, Hidekazu Shidara, Tung-Yen Chuang, and Kwan-Liu Ma. 2019.
\newblock Talktraces: Real-time capture and visualization of verbal content in meetings.
\newblock In \emph{Proceedings of the 2019 CHI conference on human factors in computing systems}, pages 1--14.

\bibitem[{Grootendorst(2022)}]{grootendorst2022bertopic}
Maarten Grootendorst. 2022.
\newblock Bertopic: Neural topic modeling with a class-based tf-idf procedure.
\newblock \emph{arXiv preprint arXiv:2203.05794}.

\bibitem[{Hopf(2004)}]{hopf2004qualitative}
Christel Hopf. 2004.
\newblock Qualitative interviews: An overview.
\newblock \emph{A companion to qualitative research}, 203(8):100093.

\bibitem[{Iijima et~al.(2021)Iijima, Shitara, Sarcar, and Ochiai}]{iijima2021word}
Ryo Iijima, Akihisa Shitara, Sayan Sarcar, and Yoichi Ochiai. 2021.
\newblock Word cloud for meeting: A visualization system for dhh people in online meetings.
\newblock In \emph{Proceedings of the 23rd International ACM SIGACCESS Conference on Computers and Accessibility}, pages 1--4.

\bibitem[{Khusro et~al.(2021)Khusro, Jabeen, and Khan}]{khusro2021tag}
Shah Khusro, Fouzia Jabeen, and Aisha Khan. 2021.
\newblock Tag clouds: past, present and future.
\newblock \emph{Proceedings of the national academy of sciences, India section A: physical sciences}, 91(2):369--381.

\bibitem[{Lennon et~al.(2021)Lennon, Fraleigh, Van~Scoy, Keshaviah, Hu, Snyder, Miller, Calo, Zgierska, and Griffin}]{lennon2021developing}
Robert~P Lennon, Robbie Fraleigh, Lauren~J Van~Scoy, Aparna Keshaviah, Xindi~C Hu, Bethany~L Snyder, Erin~L Miller, William~A Calo, Aleksandra~E Zgierska, and Christopher Griffin. 2021.
\newblock Developing and testing an automated qualitative assistant (aqua) to support qualitative analysis.
\newblock \emph{Family medicine and community health}, 9(Suppl 1):e001287.

\bibitem[{Lin et~al.(2023{\natexlab{a}})Lin, Bouneffouf, Cecchi, and Tejwani}]{lin2023neural}
Baihan Lin, Djallel Bouneffouf, Guillermo Cecchi, and Ravi Tejwani. 2023{\natexlab{a}}.
\newblock Neural topic modeling of psychotherapy sessions.
\newblock In \emph{International workshop on health intelligence}, pages 209--219. Springer.

\bibitem[{Lin et~al.(2025)Lin, Bouneffouf, Landa, Jespersen, Corcoran, and Cecchi}]{lin2025compass}
Baihan Lin, Djallel Bouneffouf, Yulia Landa, Rachel Jespersen, Cheryl Corcoran, and Guillermo Cecchi. 2025.
\newblock Compass: Computational mapping of patient-therapist alliance strategies with language modeling.
\newblock \emph{Translational Psychiatry}, 15(1):166.

\bibitem[{Lin et~al.(2023{\natexlab{b}})Lin, Cecchi, and Bouneffouf}]{lin2023psychotherapy}
Baihan Lin, Guillermo Cecchi, and Djallel Bouneffouf. 2023{\natexlab{b}}.
\newblock Psychotherapy ai companion with reinforcement learning recommendations and interpretable policy dynamics.
\newblock In \emph{Companion Proceedings of the ACM Web Conference 2023}, pages 932--939.

\bibitem[{Lin et~al.(2023{\natexlab{c}})Lin, Cecchi, and Bouneffouf}]{lin2023supervisorbot}
Baihan Lin, Guillermo Cecchi, and Djallel Bouneffouf. 2023{\natexlab{c}}.
\newblock Supervisorbot: Nlp-annotated real-time recommendations of psychotherapy treatment strategies with deep reinforcement learning.
\newblock In \emph{Proceedings of the Thirty-Second International Joint Conference on Artificial Intelligence}, pages 7149--7153.

\bibitem[{Radford et~al.(2023)Radford, Kim, Xu, Brockman, McLeavey, and Sutskever}]{radford2023robust}
Alec Radford, Jong~Wook Kim, Tao Xu, Greg Brockman, Christine McLeavey, and Ilya Sutskever. 2023.
\newblock Robust speech recognition via large-scale weak supervision.
\newblock In \emph{International conference on machine learning}, pages 28492--28518. PMLR.

\bibitem[{Rehurek et~al.(2011)Rehurek, Sojka et~al.}]{vrehuuvrek2011gensim}
Radim Rehurek, Petr Sojka, and 1 others. 2011.
\newblock Gensim—statistical semantics in python.
\newblock \emph{Retrieved from genism. org}.

\bibitem[{Rivadeneira et~al.(2007)Rivadeneira, Gruen, Muller, and Millen}]{rivadeneira2007getting}
Anna~W Rivadeneira, Daniel~M Gruen, Michael~J Muller, and David~R Millen. 2007.
\newblock Getting our head in the clouds: toward evaluation studies of tagclouds.
\newblock In \emph{Proceedings of the SIGCHI conference on Human factors in computing systems}, pages 995--998.

\bibitem[{Singh et~al.(2024)Singh, Jiang, Bhasin, Sabharwal, Moukaddam, and Patel}]{singh2024racer}
Satpreet~Harcharan Singh, Kevin Jiang, Kanchan Bhasin, Ashutosh Sabharwal, Nidal Moukaddam, and Ankit~B Patel. 2024.
\newblock Racer: An llm-powered methodology for scalable analysis of semi-structured mental health interviews.
\newblock \emph{arXiv preprint arXiv:2402.02656}.

\bibitem[{Skeppstedt et~al.(2024)Skeppstedt, Ahltorp, Kucher, and Lindstr{\"o}m}]{skeppstedt2024word}
Maria Skeppstedt, Magnus Ahltorp, Kostiantyn Kucher, and Matts Lindstr{\"o}m. 2024.
\newblock From word clouds to word rain: Revisiting the classic word cloud to visualize climate change texts.
\newblock \emph{Information Visualization}, 23(3):217--238.

\bibitem[{Steele and Iliinsky(2010)}]{steele2010beautiful}
Julie Steele and Noah Iliinsky. 2010.
\newblock \emph{Beautiful visualization: Looking at data through the eyes of experts}.
\newblock " O'Reilly Media, Inc.".

\bibitem[{Touvron et~al.(2023)Touvron, Lavril, Izacard, Martinet, Lachaux, Lacroix, Rozi{\`e}re, Goyal, Hambro, Azhar et~al.}]{touvron2023llama}
Hugo Touvron, Thibaut Lavril, Gautier Izacard, Xavier Martinet, Marie-Anne Lachaux, Timoth{\'e}e Lacroix, Baptiste Rozi{\`e}re, Naman Goyal, Eric Hambro, Faisal Azhar, and 1 others. 2023.
\newblock Llama: Open and efficient foundation language models.
\newblock \emph{arXiv preprint arXiv:2302.13971}.

\bibitem[{Wang et~al.(2020)Wang, Zhao, Guo, North, and Ramakrishnan}]{wang2020recloud}
Ji~Wang, Jian Zhao, Sheng Guo, Chris North, and Naren Ramakrishnan. 2020.
\newblock Recloud: semantics-based word cloud visualization of user reviews.
\newblock In \emph{Graphics Interface 2014}, pages 151--158. AK Peters/CRC Press.

\bibitem[{Xu et~al.(2025)Xu, Yi, Lim, Xu, Well, Mery, Zhang, Zhang, Ji, Pingali et~al.}]{xu2025tama}
Huimin Xu, Seungjun Yi, Terence Lim, Jiawei Xu, Andrew Well, Carlos Mery, Aidong Zhang, Yuji Zhang, Heng Ji, Keshav Pingali, and 1 others. 2025.
\newblock Tama: A human-ai collaborative thematic analysis framework using multi-agent llms for clinical interviews.
\newblock \emph{arXiv preprint arXiv:2503.20666}.

\end{thebibliography}

\end{document}